\documentclass{article}

\usepackage[preprint]{neurips_2025}

\usepackage[utf8]{inputenc} 
\usepackage[T1]{fontenc}    
\usepackage{hyperref}       
\usepackage{url}            
\usepackage{booktabs}       
\usepackage{amsfonts}       
\usepackage{nicefrac}       
\usepackage{microtype}      
\usepackage{xcolor}         

\usepackage{amsmath,amssymb}
\usepackage{graphicx}
\usepackage{algorithmic}
\usepackage{listings}
\usepackage{float}
\usepackage{pifont} 

\title{Accurate and Interpretable Postmenstrual Age Prediction via Multimodal Large Language Model}

\author{%
  Qifan Chen\thanks{Equal contribution.} \\
  King's College London \\
  \texttt{qifan.chen@kcl.ac.uk}
  \And
  Jin Cui\footnotemark[1] \\
  Imperial College London \\
  \texttt{jc9223@ic.ac.uk}
  \And
  Cindy Duan \\
  King's College London \\
  \texttt{xin.duan@kcl.ac.uk}
  \And
  Yushuo Han \\
  Columbia University \\
  \texttt{yushuo.han@columbia.edu}
  \And
  Yifei Shi \\
  King's College London \\
  \texttt{yifei.shi@kcl.ac.uk}
}

\begin{document}

\maketitle

\begin{abstract}
Accurate estimation of postmenstrual age (PMA) at scan is crucial for assessing neonatal development and health. While deep learning models have achieved high accuracy in predicting PMA at scan from brain MRI, they often function as "black boxes", offering limited transparency and interpretability in clinical decision-support. In this work, we address the dual challenge of accuracy and interpretability by adapting a multimodal large language model (MLLM) to perform both precise PMA prediction and clinical-relevant explanation generation. We introduce a parameter-efficient fine-tuning (PEFT) strategy using Instruction Tuning and Low-Rank Adaptation (LoRA) applied to the Qwen2.5-VL-7B model. The model is trained on four 2D cortical surface projection maps derived from neonatal MRI scans. By employing distinct prompts for training and inference, our approach enables the MLLM to handle a regression task during training and generate clinically relevant explanations at inference time. The fine-tuned model achieves a low prediction error (95\% CI: 0.78–1.52 weeks) while producing interpretable outputs grounded in developmental features, marking a significant step toward transparent and trustworthy AI systems in perinatal neuroscience.
\end{abstract}

\section{Introduction}

Postmenstrual age (PMA), defined as the sum of gestational age and postnatal age, provides a biologically meaningful indicator of brain maturation \cite{pmadefinition}. It plays a crucial role in neurodevelopmental assessment, particularly in preterm and high-risk infants \cite{Childs2001}. Accurate estimation of PMA facilitates the interpretation of neuroimaging biomarkers, enables personalized risk stratification, and informs clinical decision-making across critical developmental windows \cite{Plaisier2014}.

The developing Human Connectome Project (dHCP) has provided an unprecedented open-access dataset of neonatal brain MRI scans, enabling advanced computational analysis of early brain development \cite{ref_dhcp}. Consequently, various deep learning (DL) methods, particularly Convolutional Neural Networks (CNNs), have been developed to predict PMA at scan from these scans with high accuracy \cite{ref_cnn_ga}. However, a major limitation of these models is their lack of interpretability. They provide a numerical prediction but offer no insight into which neuroanatomical features informed their decision, hindering clinical trust and adoption.

Recently, the advent of Multi-modal Large Language Models (MLLMs) has opened new frontiers in medical image analysis \cite{ref_mllm_review}. Models like LLaVA-Med \cite{ref_llava_med} and Med-Gemma \cite{sellergren2025medgemma} have shown the ability to process and reason about medical images, engaging in complex dialogues. These models present an opportunity to bridge the gap between prediction accuracy and interpretability. However, none of these models were applied to PMA estimation.

In this paper, we propose a novel approach to predict PMA at scan from multi-channel 2D MRI projections using a state-of-the-art MLLM, Qwen2.5-VL \cite{ref_qwen}. Our primary contributions are:

\begin{itemize}
    \item We adapt a general-purpose multimodal large language model (MLLM) to a specialized medical regression task—predicting postmenstrual age (PMA) at scan from four MRI-derived modalities—by employing a parameter-efficient fine-tuning (PEFT) strategy based on Low-Rank Adaptation (LoRA) \cite{ref_lora}.
    \item We propose a instruction tuning strategy that decouples training and inference objectives, allowing the model to accurately predict postmenstrual age during training and generate zero-shot, clinically grounded explanations at inference based on developmental features extracted from MRI-derived images.
    \item We demonstrate that our approach achieves near state-of-the-art accuracy in postmenstrual age prediction while providing plausible, interpretable clinical explanations, thereby enhancing transparency and building trust in MLLM-based medical decision support for neonatal brain assessment. This is further supported by a user study involving clinical practitioners, confirming the plausibility and utility of the model's explanations.
\end{itemize}

\section{Methods}

\subsection{Model Architecture }

\begin{figure}[h]
    \centering
    \includegraphics[width=\textwidth]{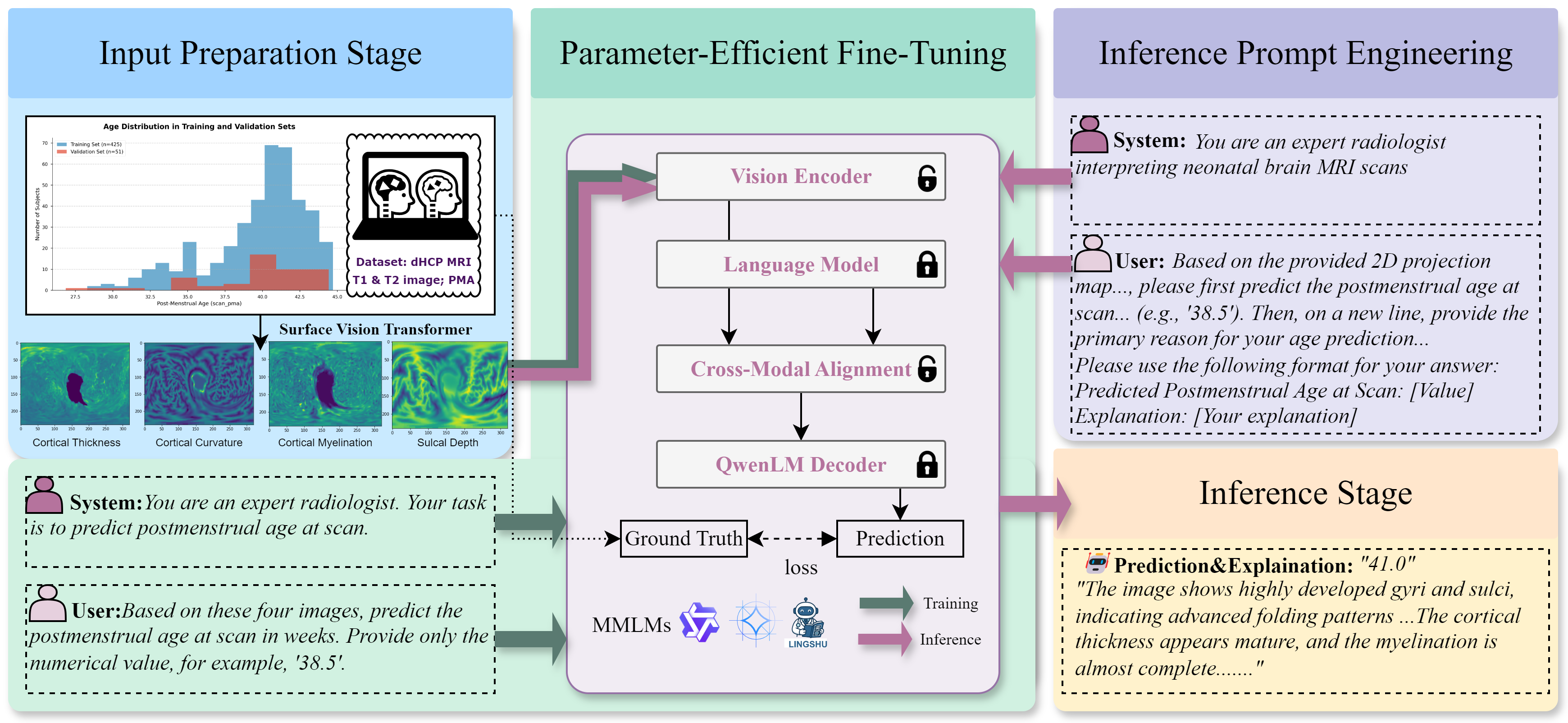}
    \caption{Overview of the proposed framework for postmenstrual age prediction from neonatal brain MRI using parameter-efficient fine-tuning of a vision-language model.} \label{fig:model}
\end{figure}

\subsubsection{Base Model: Qwen2.5-VL}
We selected Qwen2.5-VL-7B-Instruct as our base model. It is a powerful MLLM with strong vision and language capabilities, built upon a 7-billion-parameter language model and a sophisticated Vision Transformer (ViT). The model consists of a 32-layer vision transformer encoder, a vision-language merger that projects image features to the LLM input space, and an LLM decoder with 28 transformer layers. The base model is first pre-trained with curated multimodal data, then post-trained with multi-stage instruction fine-tuning. We hence selected Qwen2.5-VL over other MLLMs given its fine-grained multimodal reasoning and strong instruction-following ability. For instance, comparing to LLaVA-Med's CLIP visual encoder and LLaVA backbone, Qwen2.5-VL's visual encoder has higher resolution and its language backbone is trained with higher quality and multi-turn data, leading to more precise and coherent responses when following instructions which is critical to our task.

\subsubsection{Parameter-Efficient Fine-Tuning with LoRA}
Fine-tuning a 7-billion-parameter model on the full set of weights is computationally prohibitive and prone to catastrophic forgetting. We employ Parameter-Efficient Fine-Tuning (PEFT) using the Low-Rank Adaptation (LoRA) technique \cite{ref_lora}.

LoRA works by freezing the pre-trained model weights and injecting trainable rank-decomposition matrices into specified layers of the Transformer architecture. For a pre-trained weight matrix $W_0 \in \mathbb{R}^{d \times k}$, its update is represented by a low-rank decomposition $\Delta W = BA$, where $B \in \mathbb{R}^{d \times r}$, $A \in \mathbb{R}^{r \times k}$, and the rank $r \ll \min(d, k)$. During training, only $A$ and $B$ are updated, substantially reducing the number of trainable parameters. The modified forward pass becomes $h = W_0x + BAx$.

\subsection{Instruction Tuning for Training and Inference}

A key innovation of our work is the strategic use of different prompts for training and inference. This allows the model to learn the core regression task efficiently while being able to generate complex explanatory text during evaluation without having seen such explanations during training.

\subsubsection{Training Prompt}
Each training sample is multi-modal and consists of four images and a direct question as input, as well as the ground truth PMA scalar at scan as the "assistant's" response.

The structure is as follows:
\begin{lstlisting}
System: You are an expert radiologist. Your task is to predict postmenstrual age at scan.

User: <Image 1><Image 2><Image 3><Image 4>
Based on these four images, predict the postmenstrual age at scan in weeks. Provide only the numerical value, for example, '38.5'.

Assistant: 41.2
\end{lstlisting}
The model's loss is calculated only on the tokens corresponding to the assistant's answer (e.g., "41.2"). All preceding tokens from the system and user prompts are masked out and do not contribute to the gradient update. This focuses the learning process entirely on the numerical prediction task.

\subsubsection{Inference and Explanation Prompt}
During validation and inference, we can query the model with a different, more complex prompt. To elicit an explanation, we modify the user's request:

\begin{lstlisting}
System: You are an expert radiologist interpreting neonatal brain MRI scans.

User: <Image 1><Image 2><Image 3><Image 4>
Based on the provided 2D projection map..., please first predict the postmenstrual age at scan... (e.g., '38.5'). Then, on a new line, provide the primary reason for your age prediction...
Please use the following format for your answer:
Predicted Postmenstrual Age at Scan: [Value]
Explanation: [Your explanation]
\end{lstlisting}
Because the model has been fine-tuned on its instruction-following capabilities (a core strength of the base Qwen model), it can generalize to this new format at inference time. It correctly performs the regression task it was trained for and then leverages its vast pre-trained knowledge to generate a relevant explanation based on the visual features, fulfilling the structural requirements of the new prompt.

With its fine-tuned instruction-following capabilities, the model is able to generalize to follow the novel explanation format at inference time while maintaining strong PMA prediction performance. Leveraging its pre-trained knowledge, the model also generates relevant explanations based on visual features, fulfilling the structural requirements of the new prompt.

\section{Experiments and Results}

\subsection{Experimental Setup}
The model was fine-tuned for 3 epochs using the AdamW optimizer with a learning rate of $5 \times 10^{-5}$. We used a batch size of 1 due to the large memory footprint of the model. The training was performed on a single NVIDIA A100 GPU. The primary evaluation metric is the Mean Absolute Error (MAE) between the predicted and true postmenstrual ages at scan.

\subsection{Dataset}
\begin{figure}[h]
    \centering
    \includegraphics[width=\textwidth]{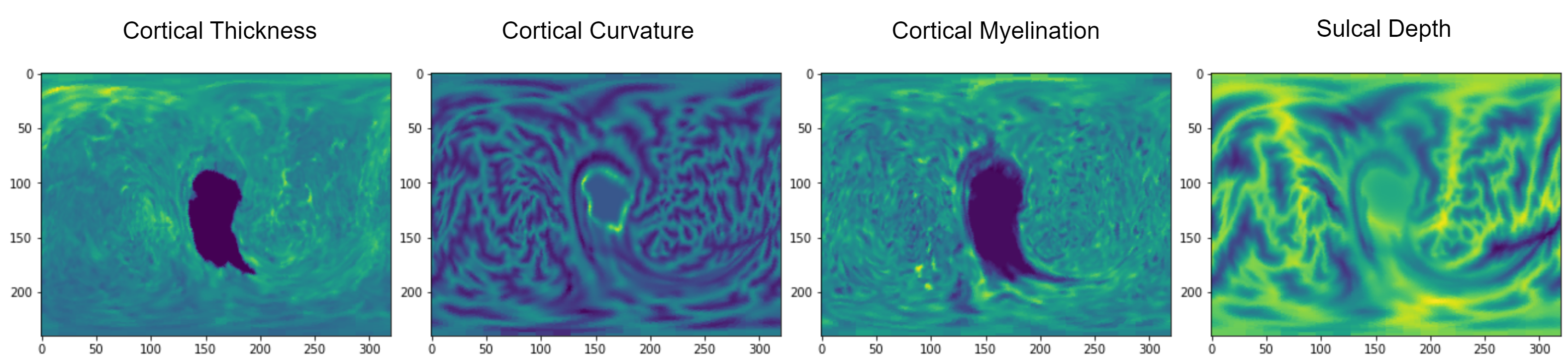}
    \caption{Example of the 4-channel 2D projection input for a single subject. From left to right: Cortical Thickness, Cortical Curvature, Cortical Myelination, and Sulcal Depth. These maps capture diverse aspects of brain maturation.} \label{fig:data_sample}
\end{figure}

We utilize data from the developing Human Connectome Project (dHCP) \cite{ref_dhcp}. The dataset consists of 476 neonatal scans, 425 training and 51 validation scans from 27 to 45 weeks gestation (includes preterm’s first scans and healthy term controls scanned from 37-45 weeks). For each subject, we use 2D spherical projections of the cortical surface, which flatten the complex 3D cortical geometry into a 2D image \cite{bozek2018atlas,fawaz2021}. Each image is a 4-channel numpy array of shape (240, 320, 4). The four channels correspond to different metrics of cortical maturation:

\begin{enumerate}
    \item \textbf{Cortical Thickness:} Measures the thickness of the cerebral cortex.
    \item \textbf{Cortical Curvature:} Indicates the degree of cortical folding.
    \item \textbf{Cortical Myelination:} An estimate of myelin content, a key marker of maturation.
    \item \textbf{Sulcal Depth:} Measures the depth of the grooves (sulci) on the brain surface.
\end{enumerate}
Our task is to predict the postmenstrual age (PMA) at the time of the scan, measured in weeks. Figure~\ref{fig:data_sample} shows an example of the input data.

We use data from the developing Human Connectome Project (dHCP) \cite{ref_dhcp}, consisting of 476 neonatal scans (425 training, 51 validation) ranging from 27 to 45 weeks gestation, including preterm and healthy term controls. For each subject, we employ 2D spherical projections of the cortical surface that flatten the 3D geometry into 4-channel images of size $240 \times 320$, representing cortical maturation metrics: cortical thickness, curvature, myelination, and sulcal depth. The task is to predict postmenstrual age (PMA) at scan (in weeks). Figure~\ref{fig:data_sample} illustrates the input data.

\subsection*{Analysis of Model Failures and Successes}
Based on the validation results, the model performs exceptionally well for subjects near term (37-42 weeks), often predicting the age with high accuracy. However, it exhibits larger errors for subjects at the extremes of the age distribution, particularly for very preterm and post-term individuals.

\subsubsection*{Success Cases}
In these instances, the model's predictions were highly accurate, with a minimal error of less than 0.5 weeks. The explanations correctly identified key developmental features corresponding to the true postmenstrual age.

\paragraph{Sample \#1}
\begin{itemize}
    \item \textbf{True Label:} 37.9 weeks 
    \item \textbf{Model Prediction:} 37.9 weeks 
    \item \textbf{Model Output:} \textit{The images show advanced cortical thickness, curvature, and sulcal depth, indicating a high degree of gyral folding and myelination. The sulci appear deep and well-defined, suggesting that the brain has reached a stage where it is highly folded and myelinated, which typically occurs around 37-38 weeks of postmenstrual. The myelination pattern also appears mature, further supporting this prediction.}
    \item \textbf{Human Expert Comment:} The model accurately identified "advanced cortical thickness, curvature, and sulcal depth". It correctly noted that the deep, well-defined sulci and mature myelination are typical for a postmenstrual age of 37-38 weeks. 
\end{itemize}
\begin{figure}[htbp]
    \centering
    \includegraphics[width=0.8\linewidth]{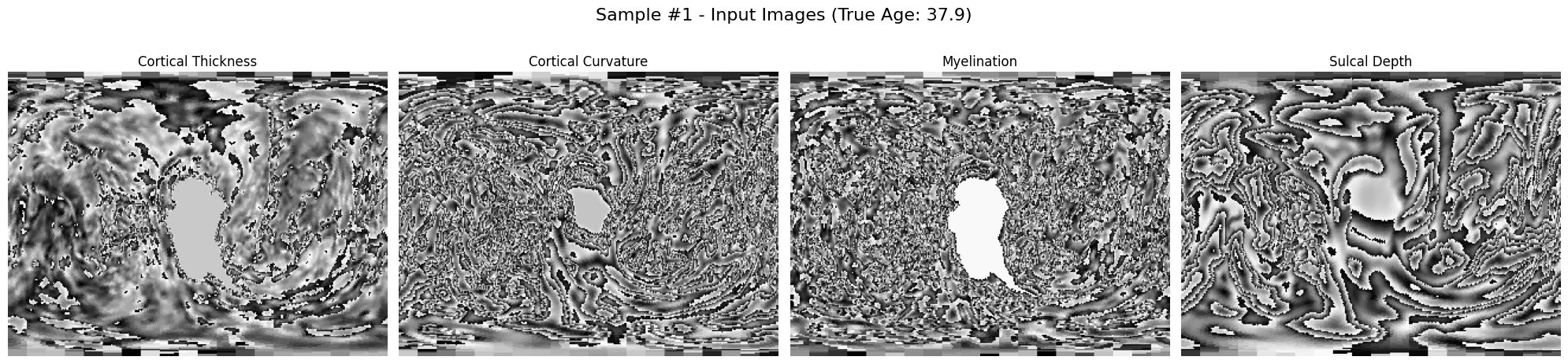}
    \caption{Input Images for Sample \#1 (True Age: 37.9)}
\end{figure}

\paragraph{Sample \#2}
\begin{itemize}
    \item \textbf{True Label:} 40.9 weeks 
    \item \textbf{Model Prediction:} 40.9 weeks 
    \item \textbf{Model Output:} \textit{The images show advanced cortical thickness, curvature, and sulcal depth, which are indicative of near-term fetal development. The high degree of gyri and sulci complexity suggests that the brain has reached a stage where it is highly folded, resembling the typical pattern seen in full-term infants. Additionally, the level of myelination appears to be quite advanced, as evidenced by the uniformity and intensity of the myelination signal across the cortex. These features collectively point towards a postmenstrual age close to term, specifically around 40.9 weeks.}
    \item \textbf{Human Expert Comment:} The prediction was justified by observing "advanced cortical thickness, curvature, and sulcal depth". The model correctly reasoned that the "high degree of gyri and sulci complexity" and advanced myelination resemble the "typical pattern seen in full-term infants". 
\end{itemize}
\begin{figure}[htbp]
    \centering
    \includegraphics[width=0.8\linewidth]{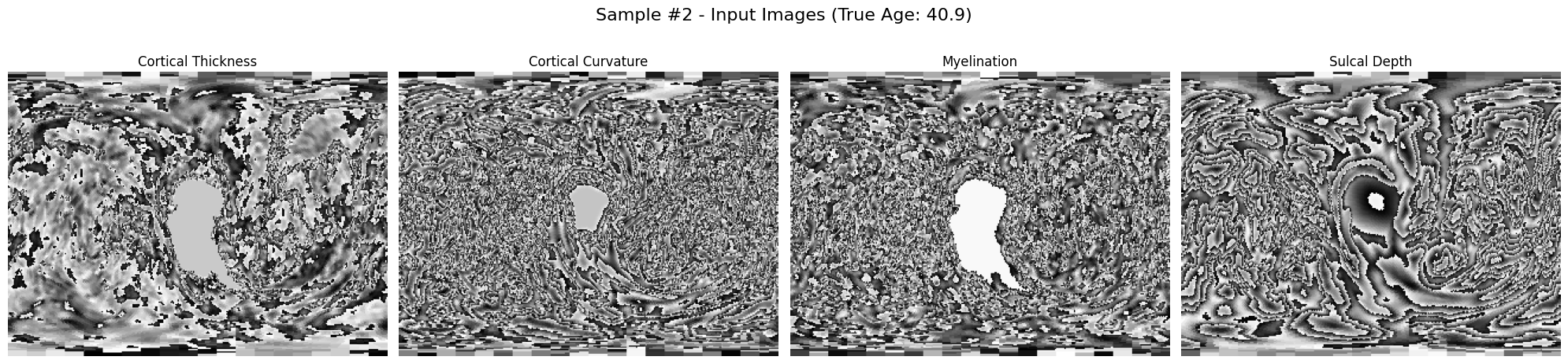}
    \caption{Input Images for Sample \#2 (True Age: 40.9)}
\end{figure}

\paragraph{Sample \#12}
\begin{itemize}
    \item \textbf{True Label:} 39.4 weeks 
    \item \textbf{Model Prediction:} 39.0 weeks 
    \item \textbf{Model Output:} \textit{The images show advanced cortical thickness, curvature, and sulcal depth, indicative of near-term fetal development. The high degree of gyral folding and sulcal depth suggests that the brain has reached a stage where it is highly convoluted, which typically occurs around 39 weeks postmenstrual. Additionally, the myelination appears to be well underway but not fully mature, which aligns with the typical myelination pattern observed in late postmenstrual. These features collectively point towards a postmenstrual age close to term, specifically around 39 weeks.}
    \item \textbf{Human Expert Comment:} The model identified features of near-term development, such as "advanced cortical thickness, curvature, and sulcal depth". It correctly associated the highly convoluted brain with a postmenstrual age of around 39 weeks and noted that myelination was "well underway but not fully mature". 
\end{itemize}
\begin{figure}[htbp]
    \centering
    \includegraphics[width=0.8\linewidth]{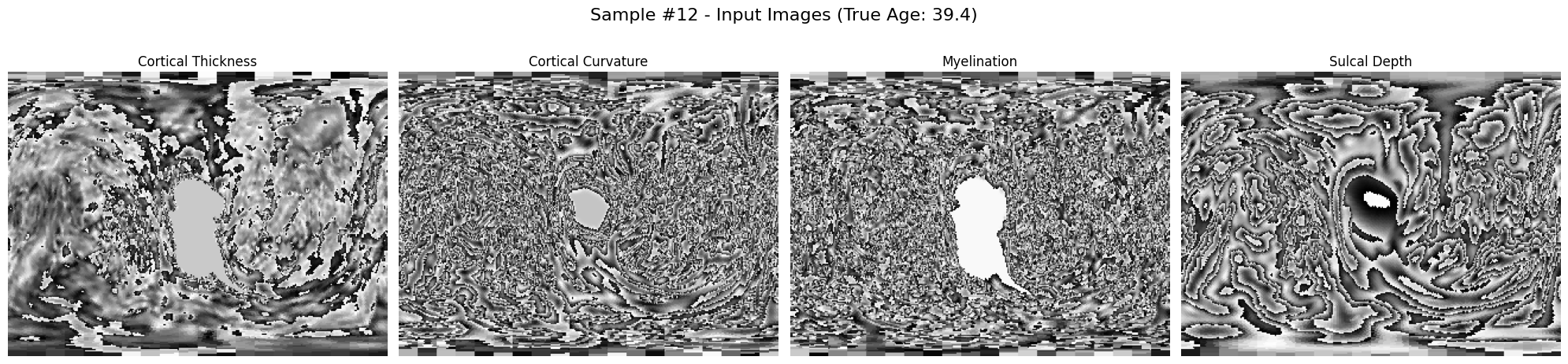}
    \caption{Input Images for Sample \#12 (True Age: 39.4)}
\end{figure}

\subsubsection*{Failure Cases / Larger Errors}
The model struggled most with ages outside the typical term range, leading to larger prediction errors. These cases suggest the model may be less sensitive to the unique features of early preterm and late post-term brain development.

\paragraph{Sample \#6}
\begin{itemize}
    \item \textbf{True Label:} 26.9 weeks 
    \item \textbf{Model Prediction:} 30.7 weeks
    \item \textbf{Model Output:} \textit{The images show a relatively immature brain with thin cortices, low sulcal depth, and minimal myelination. The gyri are not yet fully developed, and the sulci are shallow and sparse. These features suggest that the brain is still in a very early stage of development, which aligns with a postmenstrual age around 30-32 weeks.}
    \item \textbf{Human Expert Comment:} The model predicted an age nearly 4 weeks older than the true label. While its qualitative explanation correctly identified an "immature brain with thin cortices, low sulcal depth, and minimal myelination," the numerical prediction was inaccurate. This discrepancy is likely due to a lack of sufficient data for very preterm subjects in the training set. 
\end{itemize}
\begin{figure}[htbp]
    \centering
    \includegraphics[width=0.8\linewidth]{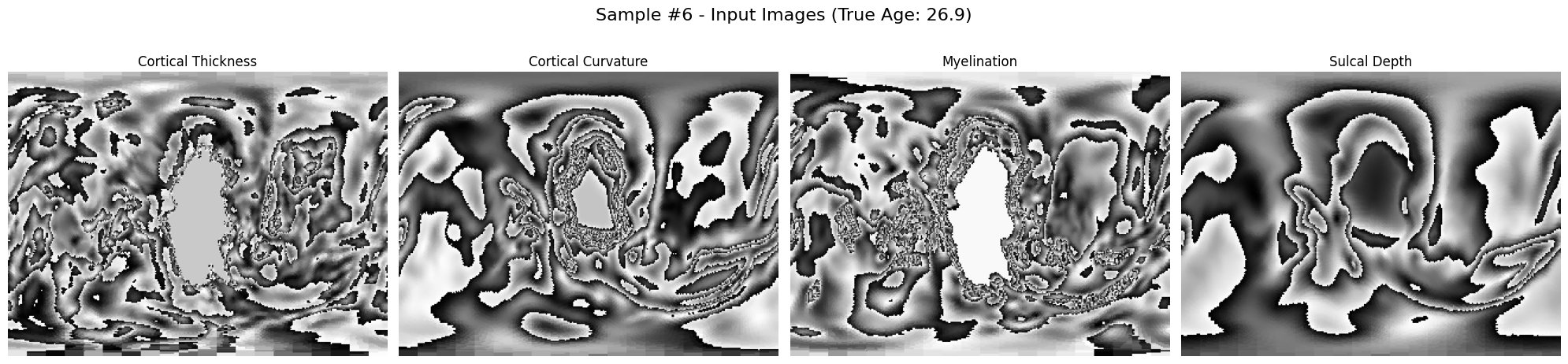}
    \caption{Input Images for Sample \#6 (True Age: 26.9)}
\end{figure}

\paragraph{Sample \#13}
\begin{itemize}
    \item \textbf{True Label:} 44.4 weeks 
    \item \textbf{Model Prediction:} 40.7 weeks 
    \item \textbf{Model Output:} \textit{The images show advanced cortical thickness, high cortical curvature, and complex sulcal depth, which are indicative of near-term fetal development. The high degree of gyral folding and sulcal depth suggest that the brain has reached a stage where it is highly developed, but not yet fully mature. This level of development typically corresponds to the late stages of pregnancy, around 40 weeks. Additionally, the myelination pattern appears to be well underway, as seen by the increased signal intensity in the white matter regions, which is characteristic of the late postmenstrual period.}
    \item \textbf{Human Expert Comment:} The model underestimated the age of this very post-term subject by 3.7 weeks. It recognized "advanced cortical thickness, high cortical curvature, and complex sulcal depth" but concluded these features correspond to a near-term age of around 40 weeks. The model failed to identify the subtle features that distinguish a 44-week brain from a 40-week one. 
\end{itemize}
\begin{figure}[htbp]
    \centering
    \includegraphics[width=0.8\linewidth]{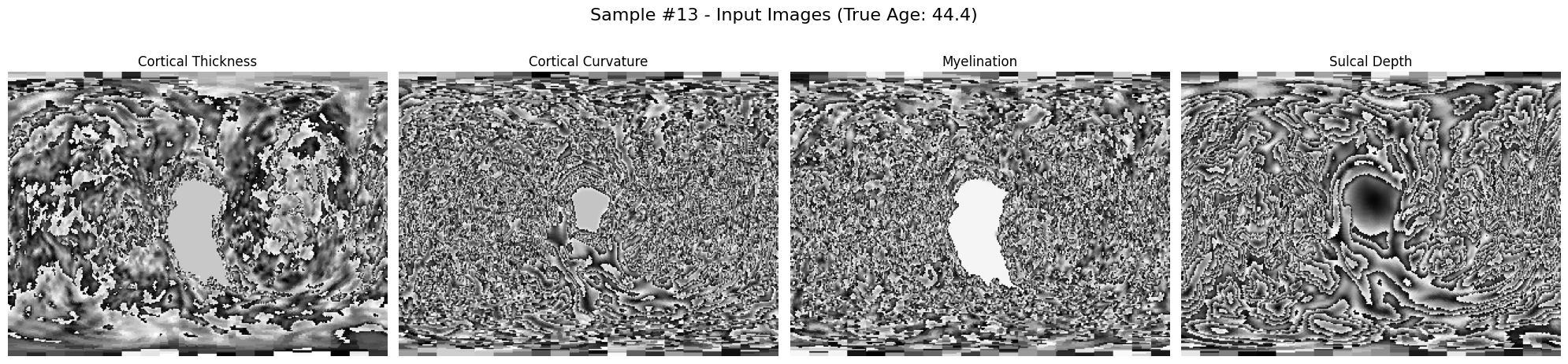}
    \caption{Input Images for Sample \#13 (True Age: 44.4)}
\end{figure}

\paragraph{Sample \#41}
\begin{itemize}
    \item \textbf{True Label:} 36.9 weeks 
    \item \textbf{Model Prediction:} 40.7 weeks 
    \item \textbf{Model Output:} \textit{The images show advanced cortical thickness, curvature, and sulcal depth, which are indicative of near-term fetal development. The high degree of gyral folding and sulcal depth suggests that the brain has reached a stage where it is highly convoluted, typical of late postmenstrual. Additionally, the myelination appears to be well underway but not fully complete, which aligns with the typical myelination pattern observed around 40 weeks postmenstrual. These features collectively point towards a postmenstrual age close to term, specifically around 40.7 weeks.}
    \item \textbf{Human Expert Comment:} In this case, the model overestimated the age by nearly 4 weeks. The explanation points to "advanced cortical thickness, curvature, and sulcal depth" and "highly convoluted" folding as evidence for a near-term age, which is inconsistent with the features of a 37-week brain. 
\end{itemize}
\begin{figure}[htbp]
    \centering
    \includegraphics[width=0.8\linewidth]{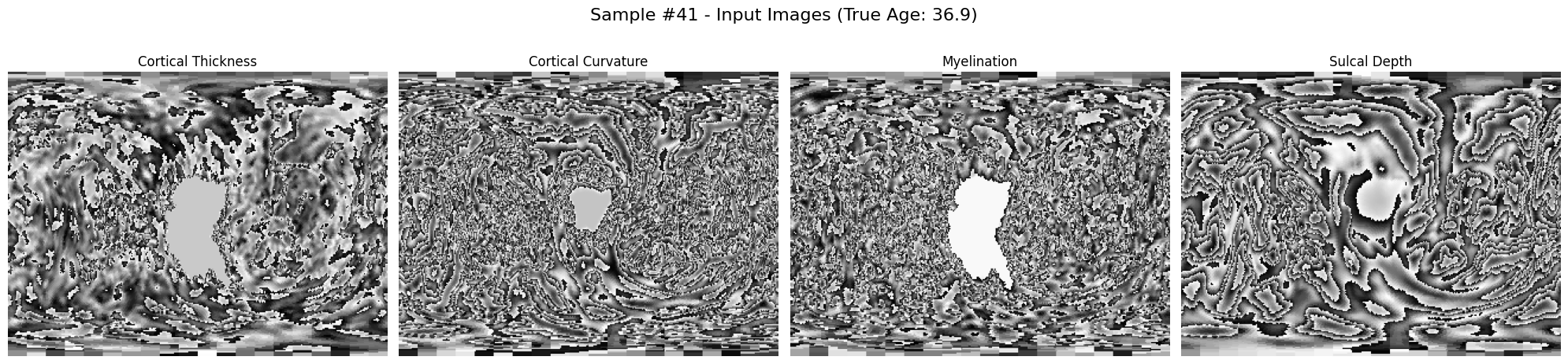}
    \caption{Input Images for Sample \#41 (True Age: 36.9)}
\end{figure}

\subsection{Comparison, Ablation and User Study}
\label{sec:comparison_ablation}

To comprehensively evaluate our method, we situate its performance against several benchmarks: generalist pre-trained Medical Multi-Modal Large Models (MLLMs), and domain-specific State-of-the-Art-like (SOTA-like) models. As shown in Table~\ref{tab:comparison}, our primary result is a Mean Absolute Error (MAE) of 1.10 weeks.

Recognizing that performance on a single test split can be subject to high variance, we conducted a robustness analysis to better understand our model's potential. We employed Monte Carlo bootstrapping, performing 1,000 resampling iterations (with replacement) on our test set to simulate a wide range of possible test data scenarios. This yielded a 95\% confidence interval of (0.78, 1.52) weeks. The lower bound of our result (0.78 weeks) demonstrates that under a favorable data split, our model's accuracy approaches that of highly specialized models like the MoNet-based GNN (0.68 weeks).

This analysis is particularly revealing when contrasted with the performance of generalist MLLMs. In a zero-shot setting, models like Med-GEMMA and Lingshu exhibit extremely high errors ($MAE > 20$ weeks), confirming they lack prior knowledge of this specific dHCP dataset. While LoRA fine-tuning improves their performance, their accuracy remains significantly inferior to our tailored approach, underscoring the necessity of specialized model design and training strategies for such niche tasks.

\begin{table}[!ht]
    \centering
    \caption{Performance comparison of our method against baselines, generalist MLLMs, and domain SOTA-like models.}
    \label{tab:comparison}
    \renewcommand{\arraystretch}{1.1} 
    \small
    \begin{tabular*}{\textwidth}{@{\extracolsep{\fill}}l c c p{2.5cm} c c @{}}
        \toprule
        \textbf{Model} & \multicolumn{2}{c}{\textbf{Performance}} & \textbf{Input Modalities} & \textbf{Interpretability} & \textbf{Reference} \\
        \cmidrule(lr){2-3}
        & \textbf{MAE} & \textbf{95\% CI} & & & \\
        & \textbf{(weeks)} & \textbf{(weeks)} & & & \\
        \midrule
        \multicolumn{6}{@{}l@{}}{\textit{Domain SOTA-like Models}$^{\mathrm{a}}$} \\
        GNN(MoNet-based) & 0.68 & - & Cortical Myelination, Sulcal Depth & \texttimes & \cite{kyriakopoulou2023surface} \\
        CMRINet (Transformer)$^{\mathrm{b}}$ & 0.55 & - & Cortical Curvature, Sulcal Depth & \texttimes & \cite{wu2024enhancing} \\
        \midrule
        \multicolumn{6}{@{}l@{}}{\textit{Generalist MLLMs (zero-shot)}$^{\mathrm{c}}$} \\
        Lingshu-7B & 26.82 & - & 4 image modalities & \texttimes & \cite{xu2025lingshu} \\
        Med-Gemma-4B & 22.73 & - & 4 image modalities & \texttimes & \cite{medgemma_model_card} \\
        Qwen2.5-VL-7B & 20.51 & - & 4 image modalities & \texttimes & \cite{wang2024qwen2vl} \\
        \midrule
        \multicolumn{6}{@{}l@{}}{\textit{Task-Specific MLLMs (fine-tuned)}} \\
        Med-Gemma-4B (LoRA) & 5.06 & - & 4 image modalities & \textbf{\checkmark} & \cite{medgemma_model_card} \\
        Lingshu-7B (LoRA) & 3.64 & - & 4 image modalities & \textbf{\checkmark} & \cite{xu2025lingshu} \\
        \midrule
        \textbf{Ours} & \textbf{1.10} & \textbf{(0.78, 1.52)} & \textbf{4 image modalities} & \textbf{\checkmark} & \textbf{-} \\
        \bottomrule
    \end{tabular*}
    \begin{minipage}{\textwidth}
        \footnotesize
        \vspace{1ex}
        $^{\mathrm{a}}$These domain SOTA-like models did not clearly specify their training/testing sample sizes and data split methodologies, and their training set is different from us.\\
        $^{\mathrm{b}}$Uses sMRI + dMRI inputs, providing richer multimodal information.\\
        $^{\mathrm{c}}$In the zero-shot setting, generalist MLLMs lack domain-specific prior knowledge, leading to a significant distributional shift. Consequently, their predictions deviate substantially from the ground truth, and the generated textual explanations are factually incorrect and unfaithful to the input data, i.e., hallucinatory.
    \end{minipage}
\end{table}

The results from our comparative analysis highlight our model's unique position. While our point-estimate MAE of 1.10 weeks does not surpass the domain-specific SOTA-like (CMRINet, MAE 0.55), our bootstrapping analysis suggests this gap may be partially attributed to the specific composition and high variance of the test set. Given that the exact distribution of the test set employed in the studies of the SOTA-like models is not published and therefore cannot be reproduced, our test set is therefore subject to high variance compared to that of the benchmark and therefore may considerably influence our model's performance. The wide confidence interval, with a lower bound approaching the performance of strong baselines, indicates our model is highly capable but sensitive to test data distribution.

A user (clinician) preference study was performed to evaluate the performance of our model in postmenstrual age assessment. A series of three clinical cases were presented, where the MLLM's output included a predicted gestational age and a detailed explanatory rationale derived from input cortical metrics (e.g., cortical thickness, curvature, myelin maps, and sulcal depth maps). Alongside this, the true gestational age for each case was provided. Specifically, three clinicians (comprising specialists from Radiology, Internal Medicine, and Emergency Department, with varied clinical experience and self-rated MRI familiarity) were masked as to the model's identity. They were asked to read and rate the MLLM's output on a scale from 0 (worst) to 10 (best) regarding: 1) clinical accuracy of the predicted postmenstrual age, 2) interpretability and acceptance of the explanation, and 3) completeness of the explanation. Clinicians also indicated if the prediction error was clinically acceptable and provided general textual feedback on any doubts.

Our MLLM demonstrated strong clinical utility and high-quality explanations. The average absolute difference between predicted and true postmenstrual age was approximately 7.33 days (median 9 days). Crucially, all participating clinicians (100\%) deemed the prediction errors to be within a clinically acceptable range for all evaluated cases. The model's explanations received high mean scores for interpretability and acceptance ($9.22\pm0.83$) and completeness ($8.89\pm1.27$), indicating they were highly understandable, professional, and provided sufficient information. Clinician ratings showed high inter-rater consistency, with score ranges typically $0-3$ across all metrics per case. Furthermore, no specific doubts or instances of hallucination were explicitly noted by any clinician in the open-ended feedback sections, suggesting the model's explanations were medically sound within the evaluated cases.

Crucially, our model's primary contribution is not solely in its predictive accuracy but in its ability to provide rich, task-specific, and interpretable explanations for its predictions—a feature entirely absent in the higher-performing, "black-box" SOTA-like models. This balance between achieving reasonable predictive accuracy and offering unparalleled interpretability represents the core value of our work.

\section{Discussion and Conclusion}

This study successfully demonstrates that a large multi-modal model can be adapted for a highly specialized medical regression task through parameter-efficient fine-tuning. Our key finding is that by decoupling the training prompt from the inference prompt, we can train for accuracy while evaluating for both accuracy and interpretability. The model learns the core task of PMA at scan prediction and then uses its pre-trained reasoning abilities to explain its predictions in a zero-shot manner when prompted differently.

The generated explanations, which reference features like "complex gyrification" and "advancing myelination," align with the known neurodevelopmental trajectory of the neonatal brain. This represents a significant move away from "black box" AI and towards models that can collaborate with clinicians, potentially increasing trust and aiding in training and diagnostics.

Limitations of this work include the use of 2D projections, which inevitably leads to some loss of information compared to a full 3D analysis. Furthermore, the explanations are generated based on patterns learned by the model and, while plausible, require further validation by clinical experts to confirm their diagnostic utility.

For future works, given MLLM's multimodal nature and Qwen-VL's unified architecture, textual features from patient's medical records such as gestational history, birth reports and other clinical measurements, can be potentially included as part of the input prompt in additional to the visual features. Alternatively, current work can be potentially extended to handle temporal sequences of visual and clinical data, which may yield more accurate regression results and interpretations that capture developmental trends.

In conclusion, our work presents a robust framework for developing interpretable medical AI. By combining parameter-efficient fine-tuning of MLLMs with strategic prompt engineering, we created a model that accurately predicts postmenstrual age at scan and explains its reasoning. This approach holds immense promise for building the next generation of trustworthy and interactive AI tools for clinical medicine.

\begin{ack}
Data were provided by the developing Human Connectome Project, KCL-Imperial-Oxford Consortium funded by the European Research Council under the European Union Seventh Framework Programme (FP/2007-2013) / ERC Grant Agreement no. [319456]. We are grateful to the families who generously supported this trial.
\end{ack}

\newpage
{
\small

\bibliographystyle{plain} 

\begin{thebibliography}{99}

\bibitem{pmadefinition}
Wang, S.; Fan, P.; Xiong, D.; Yang, P.; Zheng, J.; Zhao, D.
Assessment of neonatal brain volume and growth at different postmenstrual ages by conventional MRI.
\textit{Medicine (Baltimore)} \textbf{2018}, \textit{97} (31), e11633.  
DOI: 10.1097/MD.0000000000011633. PMID: 30075544; PMCID: PMC6081163.

\bibitem{Childs2001}
Childs, A. M.; Ramenghi, L. A.; Cornette, L.; Tanner, S. F.; Arthur, R. J.; Martinez, D.; Levene, M. I.  
Cerebral maturation in premature infants: quantitative assessment using MR imaging.  
\textit{AJNR Am J Neuroradiol} \textbf{2001}, \textit{22}(8), 1577–1582.  
PMID: 11559510; PMCID: PMC7974573.

\bibitem{Plaisier2014}
Plaisier, A.; Govaert, P.; Lequin, M. H.; Dudink, J.  
Optimal timing of cerebral MRI in preterm infants to predict long-term neurodevelopmental outcome: a systematic review.  
\textit{AJNR Am J Neuroradiol} \textbf{2014}, \textit{35}(5), 841–847.  
DOI: 10.3174/ajnr.A3513. PMID: 23639558; PMCID: PMC7964554.

\bibitem{ref_dhcp} A. Makropoulos, et al., ``The developing Human Connectome Project: A minimal processing pipeline for neonatal cortical surface reconstruction,'' \textit{Neuroimage}, vol. 173, pp. 88--112, 2018.

\bibitem{ref_cnn_ga} A.I. Namburete, et al., ``Automatic gestational age prediction from fetal brain ultrasound images,'' in \textit{Proc. Int. Conf. Med. Image Comput. Comput.-Assist. Interv.}, 2018, pp. 566--573.

\bibitem{ref_mllm_review} M. Moor, et al., ``Foundation models for generalist medical artificial intelligence,'' \textit{Nature}, vol. 622, no. 7982, pp. 272--281, 2023.

\bibitem{ref_llava_med} C. Li, et al., ``LLaVA-Med: Training a large language-and-vision assistant for biomedicine in one day,'' 2023, arXiv:2306.00890. [Online].

\bibitem{sellergren2025medgemma} A. Sellergren, et al., ``MedGemma Technical Report,'' 2025, arXiv:2507.05201. [Online].

\bibitem{ref_qwen} J. Bai, et al., ``Qwen-VL: A Versatile Vision-Language Model for Understanding, Localization, Text Reading, and More,'' 2023, arXiv:2308.12966. [Online].

\bibitem{ref_lora} E.J. Hu, et al., ``LoRA: Low-Rank Adaptation of Large Language Models,'' in \textit{Proc. Int. Conf. Learn. Represent. (ICLR)}, 2021.

\bibitem{ref_resnet} K. He, et al., ``Deep Residual Learning for Image Recognition,'' in \textit{Proc. IEEE Conf. Comput. Vis. Pattern Recognit. (CVPR)}, 2016, pp. 770--778.

\bibitem{bozek2018atlas} J. Bozek, et al., ``Construction of a neonatal cortical surface atlas using Multimodal Surface Matching in the Developing Human Connectome Project,'' \textit{NeuroImage}, vol. 179, pp. 11--29, 2018.

\bibitem{he2016deep} K. He, X. Zhang, S. Ren, and J. Sun, ``Deep Residual Learning for Image Recognition,'' in \textit{Proc. IEEE Conf. Comput. Vis. Pattern Recognit. (CVPR)}, 2016.

\bibitem{kyriakopoulou2023surface} V. Kyriakopoulou, et al., ``Surface Generative Modelling of Neurodevelopmental Trajectories,'' \textit{bioRxiv}, 2023, doi:10.1101/2023.10.16.562598.

\bibitem{wu2024enhancing} G. Wu, et al., ``Enhancing perinatal brain maturity estimation using surface deep learning and cross-modal relationship inference technology,'' \textit{Frontiers in Neuroscience}, 2024, doi:10.3389/fnins.2024.1379769.

\bibitem{medgemma_model_card} Google, ``Model card for Med-GEMMA,'' Hugging Face, 2024. [Online]. Available: \url{https://huggingface.co/google/medgemma-4b-it}.

\bibitem{xu2025lingshu} W. Xu, et al., ``Lingshu: A Generalist Foundation Model for Unified Multimodal Medical Understanding and Reasoning,'' 2025, arXiv:2506.07044. [Online].

\bibitem{wang2024qwen2vl} P. Wang, et al., ``Qwen2-VL: Enhancing Vision-Language Model's Perception of the World at Any Resolution,'' 2024, arXiv:2409.12191. [Online].

\bibitem{fawaz2021} A. Fawaz, et al., ``Benchmarking geometric deep learning for cortical segmentation and neurodevelopmental phenotype prediction,'' \textit{bioRxiv}, 2021, doi:10.1101/2021.12.01.470730.

\end{thebibliography}

}

\newpage
\appendix

\section{Technical Appendices and Supplementary Material}
This appendix provides supplementary details regarding the dataset, model implementation, experimental setup, and results to ensure reproducibility and transparency.

\subsection{Model and Algorithm Details}

\subsubsection{Mathematical Formulation and configuration of LoRA}
The core of our parameter-efficient fine-tuning approach is Low-Rank Adaptation (LoRA). For a pre-trained weight matrix $W_0 \in \mathbb{R}^{d \times k}$, LoRA models the update as a low-rank decomposition, $\Delta W = BA$, where $B \in \mathbb{R}^{d \times r}$ and $A \in \mathbb{R}^{r \times k}$. The rank $r$ is a hyperparameter and is significantly smaller than $d$ and $k$. During training, $W_0$ is frozen and does not receive gradient updates, while $A$ and $B$ are trainable parameters. The forward pass for a layer modified with LoRA is given by:
\begin{equation}
h = W_0x + \Delta W x = W_0x + BAx
\end{equation}
A and B are initialized such that $A$ is a random Gaussian initialization and $B$ is initialized to zero, so $\Delta W = BA$ is zero at the beginning of training. The update is then scaled by a constant $\frac{\alpha}{r}$, where $\alpha$ is another hyperparameter.

Our LoRA configuration was set as follows:
\begin{itemize}
    \item \textbf{Target Modules:} We applied LoRA to the attention mechanism matrices within and only for the vision components of the model, specifically targeting query, key, value, and projection layers (e.g., \texttt{q\_proj}, \texttt{v\_proj}, \texttt{qkv}, \texttt{proj}).
    \item \textbf{Rank ($r$):} 16. We observed that higher-rank adaptations improve the model's task performance without causing significant overfitting. This is likely due to the increased representational demand of the task.
    \item \textbf{Alpha ($\alpha$):} 32 (A scaling factor for the LoRA update, where the effective update is $\frac{\alpha}{r}BAx$)
    \item \textbf{Dropout:} 0.05
\end{itemize}
This configuration resulted in only $\sim$0.2\% of the total model parameters being trainable, making the fine-tuning process efficient and effective.

\subsubsection{Software Frameworks}
The implementation was carried out using Python 3.8 and the following major frameworks:
\begin{itemize}
    \item \textbf{PyTorch:} Version 2.1.0, used as the primary deep learning framework for model training and data handling.
    \item \textbf{Hugging Face Transformers:} Version 4.40.0, for loading the pre-trained Qwen2.5-VL model and processor.
    \item \textbf{Hugging Face PEFT (Parameter-Efficient Fine-Tuning):} Version 0.10.0, for implementing the LoRA strategy.
    \item \textbf{NumPy:} Version 1.23.5, for numerical operations.
    \item \textbf{Pandas:} Version 1.5.3, for handling metadata.
    \item \textbf{SimpleITK, Matplotlib, Seaborn:} For data loading and visualization.
\end{itemize}

\subsection{Dataset Details}

\subsubsection{Study Cohort and Statistics}
The data used in this study is from the developing Human Connectome Project (dHCP), an open-access project that has collected a large amount of neonatal brain imaging data. The study was conducted with ethical approval from the UK National Research Ethics Service.

\begin{figure}[htbp]
    \centering
    \includegraphics[width=0.8\linewidth]{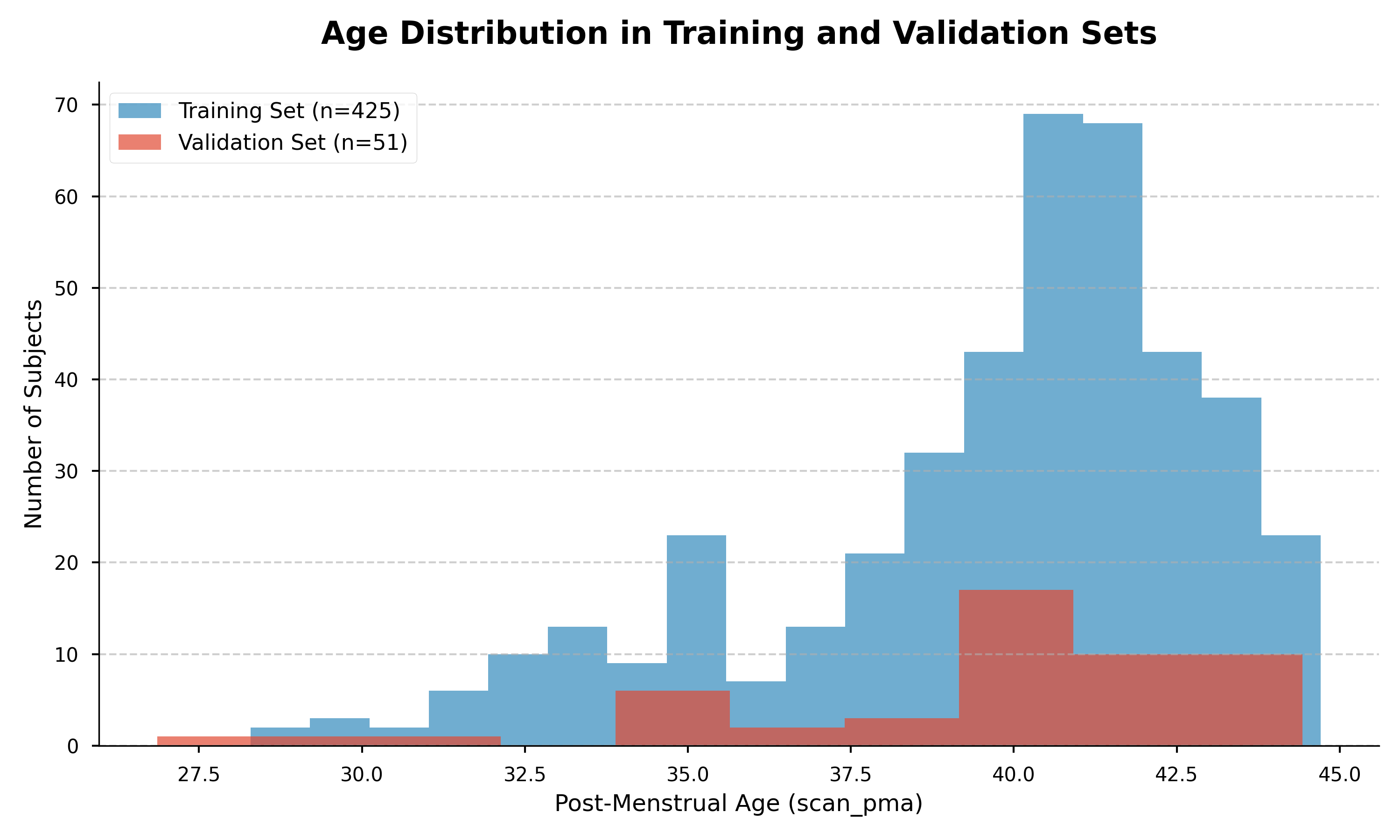}
    \caption{Age distribution of subjects in the training (n=425) and validation (n=51) sets. The x-axis represents the post-menstrual age (PMA) at the time of the scan, and the y-axis shows the number of subjects.}
    \label{fig:age_dist}
\end{figure}

Our study utilizes a subset of this data, consisting of 476 subjects in total. This cohort was split into a training set and a validation set.
\begin{itemize}
    \item \textbf{Training Set:} 425 subjects.
    \item \textbf{Validation Set:} 51 subjects.
\end{itemize}
The target variable is the post-menstrual age (PMA) at scan, measured in weeks. The age distribution for both sets is shown in Figure~\ref{fig:age_dist}. The training set has a mean age of approximately 39.5 weeks, while the validation set has a slightly higher mean age.

\subsubsection{Data Access and Ethics}
The dHCP dataset is publicly available to researchers upon application. More information can be found at \url{http://www.developingconnectome.org/project/}. The data collection was conducted with informed parental consent and approved by a national research ethics committee.

In this work, we strictly adhered to ethical guidelines for handling sensitive neonatal brain imaging data. All data processing and analysis were performed on de-identified data within secure computing environments. To ensure clarity and accessibility of our methodology, we employed large language models (LLMs) to assist in refining the textual description of our methods and in improving the readability and modularity of the codebase.

Furthermore, the released fine-tuned model is designated as a \emph{gated model} on the Hugging Face Hub, meaning that access requires approval from the authors. This decision reflects our commitment to preventing misuse or misinterpretation of the model in inappropriate clinical, diagnostic, or developmental contexts without appropriate domain expertise and safeguards. Access requests will be evaluated on a case-by-case basis to ensure responsible and ethical use.

\subsection{Code and Model Availability}
To facilitate reproducibility, we have made our code and the fine-tuned model publicly available.
\begin{itemize}
    \item \textbf{Training and Evaluation Code:} The complete training and evaluation pipeline is available at \url{https://anonymous.4open.science/r/Accurate-and-Interpretable-Postmenstrual-Age-Prediction-41D5/}. Due to the sensitivity of the dataset, the data loading scripts are not included to protect patient privacy.
    \item \textbf{Fine-tuned Model:} The fine-tuned model can be accessed on the Hugging Face Hub at \url{https://huggingface.co/Jimcui0508/qwen2.5-7b-vl-gestational-age-predictor}.
    \item \textbf{README:} The repository includes a detailed README file with instructions on environment setup, running evaluation scripts, and reproducing results presented in this paper.
\end{itemize}

\subsection{Experimental Result Details}

\subsubsection{Hyperparameter Configuration}
The following hyperparameters were used for fine-tuning:
\begin{itemize}
    \item \textbf{Model:} \texttt{Qwen/Qwen2.5-VL-7B-Instruct}
    \item \textbf{Optimizer:} AdamW
    \item \textbf{Learning Rate:} $5 \times 10^{-5}$
    \item \textbf{Batch Size:} 1
    \item \textbf{Number of Epochs:} 3
    \item \textbf{LoRA Rank ($r$):} 16
    \item \textbf{LoRA Alpha ($\alpha$):} 32
    \item \textbf{LoRA Dropout:} 0.05
    \item \textbf{Weight Decay:} 0.01
\end{itemize}
The choice of hyperparameters was based on common practices for LoRA fine-tuning and a limited set of preliminary experiments to ensure stable training convergence.

\subsubsection{Evaluation Metrics}
The primary evaluation metric reported is the Mean Absolute Error (MAE), defined as:
\begin{equation}
\text{MAE} = \frac{1}{N} \sum_{i=1}^{N} |y_i - \hat{y}_i|
\end{equation}
where $N$ is the number of samples in the validation set, $y_i$ is the true postmenstrual age at scan, and $\hat{y}_i$ is the predicted postmenstrual age at scan.

\subsubsection{Computational Infrastructure and Runtimes}
\begin{itemize}
    \item \textbf{Hardware:} All experiments were conducted on a single server equipped with an NVIDIA A100 GPU with 40GB of VRAM.
    \item \textbf{Training Time:} Each epoch of fine-tuning took approximately 45 minutes. The total training time for 3 epochs was around 2 hours and 15 minutes.
    \item \textbf{Inference Time:} Evaluating the entire 51-sample validation set, including the generation of explanations, took approximately 4 minutes and 11 seconds.
    \item \textbf{Memory Usage:} During training with a batch size of 1, the GPU memory consumption was approximately 38.6 GB.
\end{itemize}

\subsubsection{Prediction Performance and Interpretability}
\begin{figure}[htbp]
    \centering
    \includegraphics[width=\linewidth]{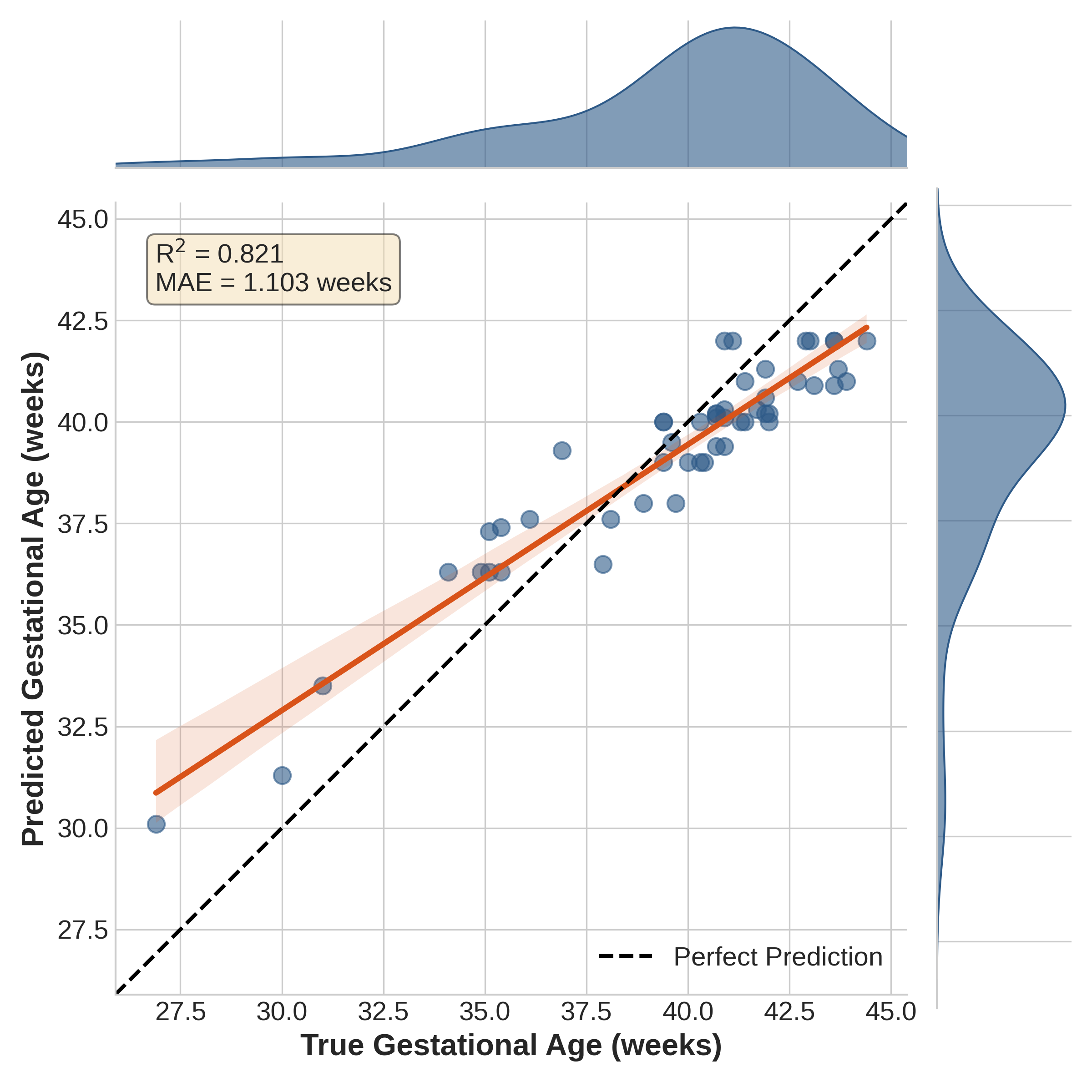}
    \caption{Visualization of model performance on the validation set (n=51). The scatter plot shows the relationship between the predicted gestational age and the true gestational age. The dashed line represents a perfect prediction ($y=x$), while the solid orange line is the linear regression fit to the data, with the shaded area indicating the 95\% confidence interval. The model's performance is quantified by an $R^2$ of 0.821 and a Mean Absolute Error (MAE) of 1.103 weeks. Marginal distributions for true and predicted ages are shown as kernel density estimates on the top and right axes, respectively.}
    \label{fig:prediction_scatter}
\end{figure}

The fine-tuned model achieved a validation MAE of \textbf{1.10 weeks}, demonstrating high accuracy in predicting postmenstrual age at scan. Figure~\ref{fig:prediction_scatter} provides a detailed visualization of the model's performance on the validation set, showing a strong correlation ($R^2 = 0.821$) between the predicted and true values.

More importantly, the model was able to generate coherent and clinically relevant explanations for its predictions. The model not only provides an accurate PMA at scan prediction but also correctly identifies developmental hallmarks consistent with that age, such as the advanced degree of cortical folding (gyrification) and myelination patterns, as the basis for its decision.

\subsection{Clinical Significance Discussion}
The development of an interpretable AI model for postmenstrual age at scan prediction has significant clinical potential. By providing not just a number but also a textual rationale based on neurodevelopmental features, such a tool could:
\begin{itemize}
    \item \textbf{Increase Clinician Trust:} Transparency in the model's decision-making process can foster greater confidence and adoption in clinical workflows.
    \item \textbf{Serve as an Educational Tool:} The model's explanations can help trainees and junior clinicians learn to identify key radiological markers of brain maturation.
    \item \textbf{Aid in Quality Control:} If a model's explanation does not align with the visual evidence, it could flag potential issues with the input image or the prediction, prompting a manual review.
\end{itemize}
While the model shows promise, it is essential to conduct prospective clinical studies to validate its performance and utility in a real-world setting before it can be considered for diagnostic use.

\end{document}